# Representing Aggregate Belief through the Competitive Equilibrium of a Securities Market

David M. Pennock and Michael P. Wellman
University of Michigan AI Laboratory
1101 Beal Avenue
Ann Arbor, MI 48109-2110 USA
{dpennock,wellman}@umich.edu

## Abstract

We consider the problem of *belief aggregation*: given a group of individual agents with probabilistic beliefs over a set of of uncertain events, formulate a sensible consensus or aggregate probability distribution over these events. Researchers have proposed many aggregation methods, although on the question of which is best the general consensus is that there is no consensus. We develop a market-based approach to this problem, where agents bet on uncertain events by buying or selling securities contingent on their outcomes. Each agent acts in the market so as to maximize expected utility at given securities prices, limited in its activity only by its own risk aversion. The equilibrium prices of goods in this market represent aggregate beliefs. For agents with constant risk aversion, we demonstrate that the aggregate probability exhibits several desirable properties, and is related to independently motivated techniques. We argue that the market-based approach provides a plausible mechanism for belief aggregation in multiagent systems, as it directly addresses self-motivated agent incentives for participation and for truthfulness, and can provide a decision-theoretic foundation for the "expert weights" often employed in centralized pooling techniques.

## 1 BELIEF AGGREGATION

The problem of *belief aggregation* is to derive a summary representation of a group's beliefs as a function of the beliefs of its constituent agents. The problem is a classic one in statistics, and it has also been well-studied in decision analysis. Despite the interest it has garnered [Benediktsson and Swain, 1992; French, 1985; Genest and Zidek, 1986; Ng and Abramson, 1992; Ng and Abramson, 1994; Weerahandi and Zidek, 1981; West, 1984], the problem has eluded definitive answers, and the plethora of impossibility results [Hylland and Zeckhauser, 1979; Genest and Zidek, 1986; Saari, 1995] and definitional controversies [French, 1985] in the literature cast doubt on the prospects for an entirely satisfactory solution.

Nevertheless, there are common situations in which a collection of agents may wish to aggregate their beliefs. As an example, a group of doctors with disparate subjective beliefs evaluating the probabilities of diseases may benefit from a formal method for combining their opinions (especially if their advice is to be culled into a probabilistic expert system [Ng and Abramson, 1992]). In more general terms, as trends toward decentralization in computation continue, aggregation mechanisms are likely to play an increasing role in uncertain reasoning. Software agents representing distinct interests and possessing individual knowledge and information-gathering capabilities will form their own beliefs, and no overarching authority will be technically or computationally able to gather all of the relevant information centrally, obtain permission to access all agents' beliefs and preferences, or enforce any globally consistent consensus. If we wish to gain the benefits of others' knowledge, we need to induce them to provide relevant reports, or perform other actions that will reveal the information we seek.

From this perspective, paramount in the design of a belief aggregation mechanism are the incentives it provides to agents to reveal their private beliefs. Given some behavioral assumptions on the participants, we aim to characterize the aggregation function "computed" by the mechanism, that is, the relationship of the derived summary to the agents' individual beliefs.

In this work, we investigate the behavior of a particular approach to belief aggregation, based on markets in uncertain propositions. The idea is that agents' decisions to trade in such markets will be driven by their beliefs and utility, and therefore the resulting prices in the markets will reflect private information bearing on the likelihood of the propositions. Agents in the market have to back up their stated positions with real money, and so have tangible disincentives to lie as well as positive incentives to participate and to gather all cost-effective relevant information.

## 2 MARKETS IN UNCERTAIN PROPOSITIONS

Our basic approach is to set up markets for uncertain propositions, essentially financial securities that pay off in monetary units contingent on uncertain events. Agents bid on these propositions according to their beliefs and preference for money. In equilibrium, the market prices can be interpreted as an aggregate probability of the participants in the market. Because each market in equilibrium balances supply and demand, the mechanism requires no subsidy, and the only cost of obtaining the aggregate belief information is that of organizing and running the markets. In other words, we do not have to pay the agents directly for revealing their information.

Let $\Omega$ denote the sample space, and capital letters near the beginning of the alphabet $(A, B, C, \ldots,$ each a proper subset of $\Omega)$, events of possible interest. Each agent has a probability distribution Pr over $\Omega$, which defines its probabilities over events of interest, $\Pr(A) = \sum_{\omega \in A} \Pr(\omega)$. With each such event $A$, we associate a *security*, worth \$1 if $A$ obtains, and zero otherwise. When the interpretation is clear by context, we may refer to the security "\$1 if $A$" simply by its corresponding event $A$.

A quantity $x$ of security $A$ can be interpreted as a lottery resulting in \$$x$ if and only if (iff) $A$ obtains. If the agent must pay a price $p$ per unit of the security, then a decision to buy $x$ units is equivalent to the lottery $L(x) = [\Pr(A), (1-p)x; -px]$, with payoff $(1-p)x$ if $A$ occurs, and $-px$ otherwise.

Let $u$ denote the agent's utility function for dollars. We can define the agent's utility, $U$, for purchasing quantities of the security, in terms of $u$,

$$\begin{aligned} U(x) &= E[u(L(x))] \\ &= \Pr(A)u((1-p)x) + (1-\Pr(A))u(-px). \end{aligned}$$

For the case of multiple securities, let $W = \{A, B, \ldots\}$ denote the set of events of interest. Let $x^Z$ represent the quantity purchased (or sold, if $x^Z < 0$) of security $Z$, for $Z \in W$, and $p^Z$ its associated price. The utility of such a bundle can also be expressed in terms of probabilities over $\Omega$ and utility for money,

$$U(x^A, x^B, \ldots) = \sum_{\omega \in \Omega} \Pr(\omega) u( \sum_{Z \in W : \omega \in Z} x^Z - \sum_{Z \in W} p^Z x^Z). \quad (1)$$

Given these beliefs and preferences, and a few behavioral assumptions, we can define the interaction of an agent in the security markets. Specifically, we assume that each agent behaves *competitively*, maximizing utility (1) at a given set of prices. We also make several additional assumptions:

1. Preferences for dollars (represented by $u$) do not depend on the uncertain state.

2. Agents do not change beliefs based on prices. That is, they do not consider that prices might reveal relevant private information of other agents.[1]

3. Agents are *risk averse* [Keeney and Raiffa, 1976; Pratt, 1964] for dollars, that is, $u$ is an increasing concave function.

The first two assumptions (along with competitive behavior) are implicit in the agent's optimization problem (1). Concavity of $u$ entails concavity of $U$, and hence any critical point of (1) is a global maximum. To ensure bounded solutions, we also generally assume $\Pr(\omega) > 0$, for all $\omega \in \Omega$. If $u$ is *sufficiently* risk averse, then the optimization problem is guaranteed to have a bounded solution for prices consistent with the logical structure of events.[2] Prices are consistent iff any combination of securities—one unit purchased or sold of each—loses money in at least one possible outcome.

The upshot of these assumptions is that each agent agrees to participate in the market and has sufficient incentive (through monetary payoffs) to make decisions according to its true beliefs. Our emphasis on explicit truth incentives is in contrast to most belief aggregation approaches, which instead merely stipulate that agents be honest. In the usual opinion pooling procedures, moreover, agents directly reveal their entire subjective probability distribution; in our model agents reveal this information only implicitly through their purchase and sale of securities.

Let the agent's *demand function*, $\mathbf{x}(\mathbf{p}) = \langle x^A(\mathbf{p}), x^B(\mathbf{p}), \ldots \rangle$, represent the quantities of securities maximizing utility (1) at prices $\mathbf{p} = \langle p^A, p^B, \ldots \rangle$.

---

[1]Since all of our results refer to equilibrium properties, these could just as well be interpreted in terms of beliefs *posterior* to information revealed through prices.

[2]Although we have not yet characterized the most general conditions, $\lim_{y \to \infty} u'(y) = 0$ is sufficient. Note that constant risk aversion, considered in detail below, has this property.



When there are multiple agents, we denote the demand, utility, and probability functions of agent $i$ by $x_i$, $u_i$, and $\Pr_i$, respectively. A market system with $N$ agents is in *competitive equilibrium* at prices **p** iff

$$\sum_{i=1}^{N} x_i(\mathbf{p}) = 0.$$

The competitive equilibrium prices in the economy represent the aggregate beliefs of the participants.

The remainder of this paper develops our market-based approach to belief aggregation in more detail. The next section considers individual behavior—how each agent determines its demand for available securities by maximizing utility. In Section 4 we examine the equilibrium prices that arise from the agents' interactions in the market. We discuss some previous proposals for market-based aggregation and other related work in Section 5, and conclude by summarizing our results and previewing future directions.

## 3  DEMAND FOR SECURITIES

The central decision facing each agent is how much of each security to demand, where positive demand indicates a quantity to buy, negative demand to sell. It acts in order to maximize expected utility, which leads to bounded behavior as long the agent is risk averse and prices are consistent. If prices are inconsistent (for example, the summed prices for a set of exhaustive events is less than one), then it is possible to identify a combination of securities that does not lose money in any possible outcome. This situation provides an opportunity for *arbitrage* [Nau and McCardle, 1991]—the portfolio can be replicated to increase utility (1) without bound. We exhibit some examples of unbounded behavior for logically related goods below.

We can derive specific conclusions about the demand behavior of agents and the resulting equilibrium prices under the assumption that agents adopt a particular form of utility function. In particular, all of our closed-form results assume that agents' preferences for money obey *constant risk aversion*, which implies that the agent's utility for dollars is given by

$$u(y) = -e^{-cy},$$

where $c$ is the agent's risk aversion coefficient [Keeney and Raiffa, 1976; Pratt, 1964].

We illustrate the agent's demand behavior for the special cases of one and two securities in Sections 3.1 and 3.2, and briefly consider the case of multiple securities in Section 3.3.

### 3.1  ONE SECURITY

Suppose an individual must decide what quantity, $x$, of a single security, "$1 if $A$", to buy or sell. The price is $p$, and the agent believes that the event will occur with probability $\Pr(A)$. The market in "$1 if $A$" is essentially a lottery $L(x)$ with payoff $(1-p)x$ if $A$ occurs, and $-px$ otherwise.

The agent's utility (1) for purchasing $x$ units of the security is the expected utility of the lottery,

$$\begin{aligned} U(x) &= E[u(L(x))] \\ &= \Pr(A)u((1-p)x) + \Pr(\bar{A})u(-px). \quad (2) \end{aligned}$$

The optimal demand must satisfy the first-order condition $U'(x) = 0$, yielding:

$$\frac{\Pr(A)}{\Pr(\bar{A})} u'((1-p)x) = \frac{p}{(1-p)} u'(-px). \quad (3)$$

**Proposition 1 (Qualitative Single-Good Demand)** *A risk-averse agent's demand for a single security is positive (zero, negative) if its probability for the corresponding event is greater than (equal to, less than) the security's price.*

**Proof.** From the first-order condition (3), $\Pr(A) > p$ implies $u'(-px) > u'((1-p)x)$. Since risk aversion entails decreasing marginal utility, the latter inequality holds iff $x > 0$. Analogous arguments establish the zero and negative-demand cases. $\square$

For agents with constant risk aversion, the utility (2) becomes

$$U(x) = E[u(L(x))] = -\Pr(A)e^{-c(1-p)x} - \Pr(\bar{A})e^{cpx}.$$

In this instance, solving the first-order condition (3) yields a closed form for the optimal demand,

$$x = \frac{1}{c} \ln \left( \frac{(1-p)}{p} \frac{\Pr(A)}{\Pr(\bar{A})} \right). \quad (4)$$

As Proposition 1 dictates, the agent's demand is directly related to its belief in the probability of $A$, and inversely related to the price of the good. At $p = \Pr(A)$, its demand is zero.[3] As an agent's risk aversion approaches zero (approximating risk neutrality) it is willing to buy or sell increasing amounts of the good, assuming the price is not equal to its belief. In a sense, a smaller risk aversion indicates greater

---

[3] If we relax our assumption that $\Pr(\omega) > 0$, demand may become unbounded. This poses a computational problem, but does *not* violate rationality—if you are *absolutely* certain of the outcome of an event you should be willing to bet arbitrarily large amounts on that eventuality, independent of risk aversion.



confidence in beliefs; the agent is willing to put more on the line when it thinks the price is too low or high. As demonstrated below, risk aversion plays the role in our model that "expert weights" play in common pooling procedures, encoding some sort of confidence, reliability, or importance factor for each individual.

## 3.2 TWO SECURITIES

Next we consider the slightly more general case of two uncertain events $A$ and $B$, and two corresponding securities "$1 if $A$" and "$1 if $B$". The going prices are $p^A$ and $p^B$, and the agent's demands are $x^A$ and $x^B$. The market in these two goods is essentially a lottery $L(x^A, x^B)$ with a payoff depending on how much the agent purchases at what price, and on the outcome of the events $A$ and $B$. Following the progression of Section 3.1, the utility for the two securities is the expected utility of this lottery, which for constant risk aversion is given by

$$\begin{aligned}
U(x^A, x^B) &= E[u(L(x^A, x^B))] = \\
&- \Pr(AB) e^{-c\left[(1-p^A)x^A + (1-p^B)x^B\right]} \\
&- \Pr(A\bar{B}) e^{-c\left[(1-p^A)x^A - p^B x^B\right]} \\
&- \Pr(\bar{A}B) e^{-c\left[-p^A x^A + (1-p^B)x^B\right]} \\
&- \Pr(\bar{A}\bar{B}) e^{-c\left[-p^A x^A - p^B x^B\right]}.
\end{aligned} \quad (5)$$

The decision variables in this optimization problem are coupled; optimal demand for $A$ may depend on the demand for $B$, and thus on the price of $B$. Figure 1 graphs utility (5) as a function of $x^A$ and $x^B$ for a particular instantiation of beliefs, risk coefficient, and prices. Although we do not have a closed form for optimal demand (except in special cases; see the treatment of mutually exclusive events below, and also Proposition 2), the problem is solvable numerically. The utility function (5) is unimodal, and optimization techniques such as Newton's method are well behaved. We can also implicitly represent the maximum of (5) as follows:

$$x^A = \\ \frac{1}{c} \ln\left(\frac{(1-p^A)}{p^A} \frac{\left[\Pr(AB)e^{-cx^B} + \Pr(A\bar{B})\right]}{\left[\Pr(\bar{A}B)e^{-cx^B} + \Pr(\bar{A}\bar{B})\right]}\right) \quad (6)$$

$$x^B = \\ \frac{1}{c} \ln\left(\frac{(1-p^B)}{p^B} \frac{\left[\Pr(AB)e^{-cx^A} + \Pr(\bar{A}B)\right]}{\left[\Pr(A\bar{B})e^{-cx^A} + \Pr(\bar{A}\bar{B})\right]}\right) \quad (7)$$

Using these equations we can numerically compute the optimum by starting at $\{x^A, x^B\} = \{0, 0\}$ and iteratively plugging the results back into the equations, until the process converges to a desired accuracy. The

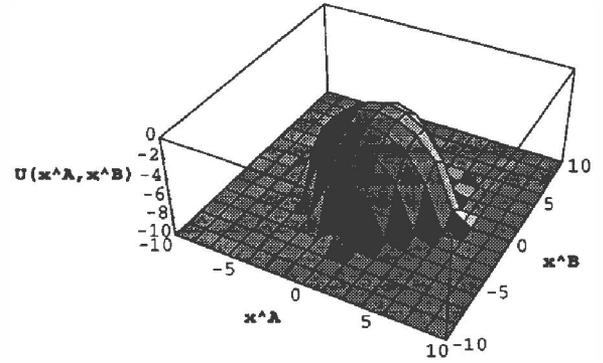

Figure 1: Utility (5) versus $x^A$ and $x^B$ for the parameters $\Pr(AB) = \Pr(A\bar{B}) = \Pr(\bar{A}B) = \Pr(\bar{A}\bar{B}) = 0.25$, $p^A = p^B = 0.5$, and $c = 1$. Utility is a maximum at $\{x^A, x^B\} = \{0, 0\}$, as expected when price equals belief.

above implicit form is also useful for examining in more detail some of the properties of an agent's demand behavior for the case of two goods.

**Proposition 2 (Separation of Independent Events)** *If the agent believes that the two events $A$ and $B$ are independent, then the optimization problem is separable with respect to $x^A$ and $x^B$, and each reduces to the single-good case (4).*

**Proof.** If the agent believes that $A$ and $B$ are independent, then (6) becomes

$$\begin{aligned}
x^A &= \\
&\frac{1}{c} \ln\left(\frac{(1-p^A)}{p^A} \frac{\Pr(A)\left[\Pr(B)e^{-cx^B} + \Pr(\bar{B})\right]}{\Pr(\bar{A})\left[\Pr(B)e^{-cx^B} + \Pr(\bar{B})\right]}\right) \\
&= \frac{1}{c} \ln\left(\frac{(1-p^A)}{p^A} \frac{\Pr(A)}{\Pr(\bar{A})}\right),
\end{aligned}$$

and similarly for $x^B$. □

When the events are dependent, demand for the two goods are correlated in the opposite direction of the dependence.

**Proposition 3 (Demand Correlation for Dependent Events)** *Demand $x^A$ is increasing in $x^B$ when $\Pr(A|B) < \Pr(A|\bar{B})$. The opposite relation holds when $\Pr(A|B) > \Pr(A|\bar{B})$.*

**Proof.** By examining (6) and (7), we see that $x^A$ and $x^B$ are positively correlated when

$$\frac{\Pr(AB)}{\Pr(\bar{A}B)} < \frac{\Pr(A\bar{B})}{\Pr(\bar{A}\bar{B})}$$

$$\frac{\Pr(A|B)}{\Pr(\bar{A}|B)} < \frac{\Pr(A|\bar{B})}{\Pr(\bar{A}|\bar{B})}$$

$$\Pr(A|B) < \Pr(A|\bar{B}).$$



The proof of the condition for negative correlation is analogous. Proposition 2 establishes zero correlation in the boundary case. □

One way to interpret this result is that negatively correlated goods provide *insurance* for each other, whereas positively related events increase the exposed risk. Put another way, positively related goods are (partial) *substitutes*, whereas negatively related goods are *complementary*. Demand correlations occur because both securities pay off in a common currency (dollars) for which agents are risk-averse.

The extreme case of negative correlation (complementarity) is disjoint events. For a pair of mutually exclusive but not exhaustive securities, we can solve (6) and (7) in closed form.

$$x^A = \frac{1}{c}\ln\left(\frac{(1-p^A-p^B)\Pr(A)}{p^A(1-\Pr(A)-\Pr(B))}\right), \quad (8)$$

and similarly for $x^B$. The solution assumes $p^A + p^B < 1$, for if this were not the case, the agent could achieve unbounded utility by selling infinite amounts of both securities. Such opportunities are always possible for logically related events, if the prices are not consistent.

For example, consider the case of two securities representing equivalent events, $A = B$. If $p^A \neq p^B$, then the path to unbounded utility is to buy infinite amounts of the cheaper security to sell in the higher priced market. If the prices coincide, then the agent acts as if there were a single security, splitting demand arbitrarily.

**Proposition 4 (Equivalent Events)** *Suppose $A = B$ and $p^A = p^B$. Then the sum of optimal demands, $x^A + x^B$, equals the demand for the single good $A$ at price $p^A$.*

**Proof.** If $A = B$, then (6) becomes:

$$\begin{aligned} x^A &= \frac{1}{c}\ln\left(\frac{(1-p^A)}{p^A}\frac{\Pr(A)e^{-cx^B}}{\Pr(\bar{A})}\right) \\ &= \frac{1}{c}\ln\left(\frac{(1-p^A)}{p^A}\frac{\Pr(A)}{\Pr(\bar{A})}\right) - x^B \\ &= D_A - x^B, \end{aligned}$$

where $D_A$ is the single-good demand for $A$ at price $p^A$ (4). Similarly, $x^B = D_B - x^A$. If $p^A = p^B$, then $x^A + x^B = D_A = D_B$. □

Figures 2 and 3 graph expected utility (5) versus $x^A$ and $x^B$ for the above situation, $A = B$. The former illustrates the case when $p^A = p^B$, the latter when $p^A > p^B$.

**Proposition 5 (Complementary Events)** *Suppose $A = \bar{B}$. If $p^A + p^B = 1$, then $x^A - x^B$ equals the demand for the single good $A$ at price $p^A$.*

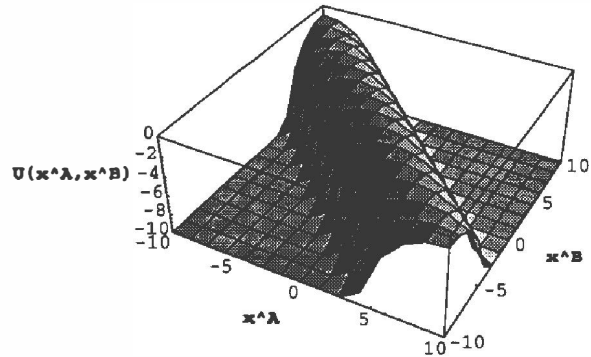

Figure 2: Utility (5) versus $x^A$ and $x^B$ for the parameters $\Pr(AB) = \Pr(\bar{A}\bar{B}) = 0.5$, $\Pr(A\bar{B}) = \Pr(\bar{A}B) = 0$, $p^A = p^B = 0.5$ and $c = 1$. In this case $A = B$. Utility is a maximum along the line $x^A + x^B = 0$; the agent's total demand for $A = B$ is zero, but it will split its purchases arbitrarily between the equivalent markets.

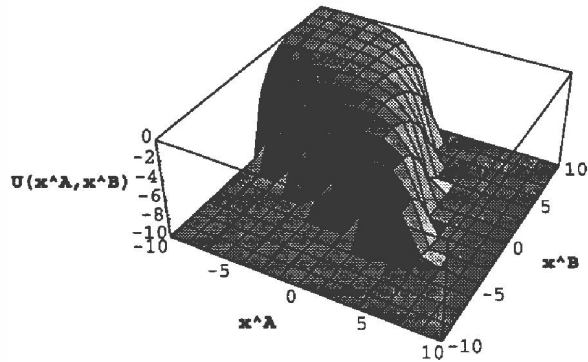

Figure 3: Utility (5) versus $x^A$ and $x^B$ for the parameters $\Pr(AB) = \Pr(\bar{A}\bar{B}) = 0.5$, $\Pr(A\bar{B}) = \Pr(\bar{A}B) = 0$, $p^A = 0.7$, $p^B = 0.3$, and $c = 1$. The prices are inconsistent with the logical relationship $A = B$, and the agent maximizes utility by selling $A$ and buying $B$ in infinite quantities.

**Proof.** Analogous to Proposition 4. □

Note that prices inconsistent with the logical structure of events cannot be part of an equilibrium, as the agents will choose infinite demands. Indeed, if any agent believes this to be the case (i.e., the agent takes some primitive events to be impossible, while others do not), then that agent will effectively dictate such a relationship in the aggregate probability.

### 3.3 MULTIPLE SECURITIES

The basic principles developed above for the one- and two-good case carry over to larger numbers of events and their corresponding securities. Any set of events that are independent from the remaining events may be handled separately. In general, for dependent

events, the demands will be correlated. The properties of logically related securities remain valid.

It is a straightforward task to write down each agent's expected utility for multiple securities, although the number of terms may be exponential in the number of events. Since the utility function is concave in securities, simple numeric maximization methods such as gradient descent or Newton's method should work well.

## 4 EQUILIBRIUM PRICES AND CONSENSUS BELIEFS

We next examine how the collective decisions of a group of agents affect the market's equilibrium prices. These prices are interpreted as the group consensus or aggregate belief in the probabilities of the associated uncertain events. In Section 4.1 we investigate in detail the case of one security, and obtain some closed-form results. In Section 4.2, we discuss the general situation of multiple events and securities.

### 4.1 ONE SECURITY

When a group of $N$ agents with constant risk aversion bid on a single security, "\$1 if $A$", we can solve directly in closed form for the competitive equilibrium price. The market is in equilibrium when $\sum_{i=1}^{N} x_i(p) = 0$, where each $x_i(p)$ is the single good demand function (4) for agent $i$. Solving this equation for $p$, we obtain

$$p = \frac{\prod_{i=1}^{N}[\Pr_i(A)]^{\alpha_i}}{\prod_{i=1}^{N}[\Pr_i(A)]^{\alpha_i} + \prod_{i=1}^{N}[\Pr_i(\bar{A})]^{\alpha_i}}, \qquad (9)$$

where $\alpha_i = (1/c_i)/\sum_j (1/c_j)$. $\Pr_i$ and $c_i$ represent the beliefs and risk aversion, respectively, of agent $i$.

Perhaps surprisingly, the competitive equilibrium price (9) turns out to be a normalized form of the *logarithmic opinion pool* (LogOP) for a single event. This standard form (essentially a geometric mean) has appeared prominently in the belief-aggregation literature [Dalkey, 1975; French, 1985; Weerahandi and Zidek, 1981; West, 1984]. Our market model, therefore, provides one way to ground a well-known *centralized* pooling mechanism in terms of individual behavior.

It also provides a decision-theoretic interpretation for the notoriously slippery concept of "expert weights". In the usual interpretation, the exponents in (9) encode some sort of degree of expertise, confidence, or reliability, and are almost always chosen in an ad hoc manner [Benediktsson and Swain, 1992; French, 1985; Genest and Zidek, 1986; Ng and Abramson, 1994]. In arguing against the use of opinion pools, French [1985] presents as his first reason that

> ...they all introduce weights ... which are not operationally defined. How should they be chosen? To say that one expert is twice as good as another is a figure of speech, not an arithmetic statement. The problem is further complicated by the likelihood of correlation between the expert's [sic] opinions. Several pragmatic solutions have been proposed ..., but to my knowledge none avoid a certain arbitrariness.

In our model, the *derived* weights are a normalized inverse measure of risk aversion, $\alpha_i = (1/c_i)/\sum_j (1/c_j)$. Note also that the weights sum to unity, as is the standard convention. Finally, allowing agents with other forms of risk aversion (other than constant) suggests a natural way to generalize the normalized LogOP.

The equilibrium price for a single security has several desirable or otherwise interesting properties as an aggregate assessment of a probability. We list them here without proof, as they all follow simply or have been observed elsewhere in the context of LogOPs.

- $\min \Pr_i(A) \le p \le \max \Pr_i(A)$

- If all agents agree that $\Pr_i(A) = k$, then $p = k$. Note that this property follows from the previous.

- If $\Pr_i(A)$ increases (decreases) and $\Pr_j(A)$ for $j \ne i$ remains fixed, then $p$ increases (decreases).

- As $c_i \to 0$ (agent $i$ becomes more risk neutral or more confident) and $c_j$ remains a positive constant for $j \ne i$, $p \to \Pr_i(A)$.

- If all risk aversion coefficients are multiplied by a positive constant, $p$ does not change.

- If a set of agent beliefs $\Pr_i(A)$ leads to an equilibrium price $p*$, then the set of agent beliefs $1 - \Pr_i(A)$ leads to an equilibrium price $1 - p*$.

Often in belief aggregation work it is desirable to show that the group as a whole behaves rationally in some sense. The following proposition establishes such a characteristic in our system, for the case of a single event.

**Proposition 6 (Equivalent "super-agent")** *Suppose a group of $N$ agents with beliefs $\Pr_1(A), \ldots, \Pr_N(A)$ and risk aversion coefficients $c_1, \ldots, c_N$ effect an equilibrium price $p*$. Their aggregate (total) demand is equal to the demand of a single*



*representative "super-agent" with belief* $\Pr(A) = p*$ *and risk aversion $c$ such that* $1/c = (1/c_1+\ldots+1/c_N)$.

**Proof.** From (4), the super-agent's demand is:

$$x = \frac{1}{c} \ln \left( \frac{(1-p)}{p} \frac{p*}{(1-p*)} \right).$$

Substituting the RHS of (9) for the equilibrium price p*, we get:

$$\begin{aligned}
x &= \frac{1}{c} \ln \left( \frac{(1-p)}{p} \frac{\prod_{i=1}^{N}[\Pr_i(A)]^{\frac{c}{c_i}}}{\prod_{i=1}^{N}[\Pr_i(\bar{A})]^{\frac{c}{c_i}}} \right) \\
&= \left( \frac{1}{c_1} + \cdots + \frac{1}{c_N} \right) \ln \left( \frac{1-p}{p} \right) + \\
&\quad \frac{1}{c_1} \ln \left( \frac{\Pr_1(A)}{\Pr_1(\bar{A})} \right) + \cdots + \frac{1}{c_N} \ln \left( \frac{\Pr_N(A)}{\Pr_N(\bar{A})} \right) \\
&= \frac{1}{c_1} \ln \left( \frac{(1-p)}{p} \frac{\Pr_1(A)}{\Pr_1(\bar{A})} \right) + \cdots + \\
&\quad \frac{1}{c_N} \ln \left( \frac{(1-p)}{p} \frac{\Pr_N(A)}{\Pr_N(\bar{A})} \right).
\end{aligned}$$

□

To an outside observer, this super-agent behaves as a rational "individual" in exactly the same sense as delineated for single agents in Section 3—the aggregate behavior can be rationalized as maximization of expected utility (with constant risk aversion, at that). In the same way, any subset of the agents can be aggregated according to Proposition 6, interacting with the rest of the system in a manner indistinguishable from an individual's behavior. Moreover, for any group of agents (assuming finite positive $c_i$) the super-agent risk aversion is *strictly less than* that of any individual. Thus the group as a whole is willing to take on more risk, acting in some sense with more "confidence" than any member alone.

### 4.2 MULTIPLE SECURITIES

The analysis is more difficult for markets in multiple securities. In general, the price of one good will depend on the prices of other goods, due to agents' correlated demand (Proposition 3). As a first step toward a more general treatment, we have derived a closed-form solution for the equilibrium security prices for two disjoint events. This involves solving for $p^A$ and $p^B$ such that $\sum_i x_i^A(p^A, p^B) = 0$ and $\sum_i x_i^B(p^A, p^B) = 0$, where $x_i^A(p^A, p^B)$ is given by (8) and $x_i^B(p^A, p^B)$ is analogous. The solution for $p^A$ is:

$$p^A = \frac{\prod_{i=1}^{N}[\Pr_i(A)]^{\alpha_i}}{\prod_{i=1}^{N}[\Pr_i(A)]^{\alpha_i} + \prod_{i=1}^{N}[\Pr_i(B)]^{\alpha_i} + \prod_{i=1}^{N}[\Pr_i(\bar{A}\bar{B})]^{\alpha_i}},$$

and similarly for $p^B$, where $\alpha_i = (1/c_i)/\sum_j(1/c_j)$. Once again we arrive at the normalized LogOP.

Our ongoing work is investigating the general cases of multiple securities, disjoint or not. It is straightforward enough to set up markets for the general case, and define their equilibria. However, except for the independent case (Proposition 2), we have not yet characterized the results for such markets.

## 5   RELATED WORK

The idea of using markets as a belief aggregation mechanism was proposed as early as forty years ago by Eisenberg and Gale [1959]. Inspired by the common method for deriving odds in horse races, they consider a pari-mutuel scheme where agents place bets across a partition of events, yielding a "consensus probability" equal to the proportion of the total bet on each event. If event $H$ obtains, agents share the total amount bet in proportion to their bets on $H$. Agents bet to maximize expected payoffs, where expectation is with respect to their probability distribution over the events, subject to a budget constraint limiting their total bets. Eisenberg and Gale show that this mechanism yields a unique set of equilibrium probabilities, and Norvig [1967] presents a dynamic process for reaching this equilibrium through iterated bids.

As the authors point out, however, this scheme can yield pathological (their word) results. For example, if there are two bettors with equal budgets, then whichever has more uniform probabilities will dictate the results. According to Genest and Zidek [1986], the pari-mutuel approach to belief aggregation "has never enjoyed much popularity" for this reason.

The pathological behavior, we believe, can be attributed to the role of arbitrary budgets. In the approach developed here, we impose no budgets, but rather rely on risk aversion to limit bets to the finite range.

Another exploration of markets for belief aggregation is Hanson's proposal for "Idea Futures" markets to encourage honest revelation of beliefs about future developments in science, technology, or other arenas of public interest [Hanson, 1995]. Although the mechanism has not been analyzed as a formal protocol, it has been operational as a game (i.e., no real money) on the world-wide web since 1994. The current version is called the "Foresight Exchange", operating at http://www.ideosphere.com/. Hanson has also described a scenario for employing a similar mechanism for coordinating computational agents [Hanson, 1991].

A third popular example of market-based belief aggregation is the Iowa Electronic Markets (IEM) sys-



tem (http://www.biz.uiowa.edu/iem/), which runs real-money markets in uncertain political and financial propositions. Their market in the US Presidential election, for example, attracted wide participation and following. *Slate* magazine even used IEM as its main index of the election "horse race".

In our previous work, we defined a general mapping from Bayesian networks to computational economies [Pennock and Wellman, 1996], showing that any joint probability distribution can be expressed as the competitive equilibrium of a particular configuration of consumer and producer agents for a particular set of securities. In this model (called *MarketBayes*), consumers have preferences directly over securities. The model presented here takes a more fundamental approach, deriving preferences over securities from underlying utility for money and beliefs in the probability of events. In addition, the analysis here focuses on aggregation of beliefs, whereas the initial MarketBayes system—though built with aggregation in mind—expressly employed only one agent with particular beliefs about each conditional event.

Although most work in economics and finance on securities markets assumes (or entails) homogeneous beliefs, Varian [1989] presents a model where agents have divergent probability distributions. He considers the effects of belief dispersion on asset valuation and trading volume, under various preference models.

The majority of work on belief aggregation does not involve markets, focusing instead on centralized mathematical pooling functions that take as input all agents' probability distributions and return as output a consensus distribution [Benediktsson and Swain, 1992; French, 1985; Genest and Zidek, 1986; Ng and Abramson, 1992; Ng and Abramson, 1994; Weerahandi and Zidek, 1981; West, 1984]. Typically these models do not consider interaction protocols, incentives for truth or for participation, or how to assign "expert weights" to the individuals involved. Researchers have debated desirable properties of aggregation schemes, producing axiomatic arguments for some, including the standard linear and logarithmic (LogOP) opinion pools.

In particular, West [1984] defines $S_i$ as the value for which individual $i$ is indifferent between $D_i$ units of security $A$, and $S_i$ dollars for sure. From this and $u_i$, one can derive $\Pr_i(A)$. Consider an assumption that the group as a whole will be indifferent between the allocation of $D = \langle D_1, \ldots, D_N \rangle$ units of the security, and the certain payoff $S = \langle S_1, \ldots, S_N \rangle$. From the premise that this is true for any $D$ (along with some other conditions), West proves that the group's $\Pr(A)$ must obey a geometric mean:

$$\Pr(A) = \prod_{i=1}^{N} [\Pr_i(A)]^{w_i},$$

which is the *unnormalized* LogOP. In contrast to the normalized version (9), this pooling function does *not* in general yield a probability (e.g., if any agents disagree, $\Pr(A) + \Pr(\bar{A}) < 1$). West goes on to show that the degree by which the aggregate diverges from a probability can be used as a measure of disagreement among the agents.[4]

However, we would make the case that the group should not be indifferent between $S$ dollars and security distribution $D$, unless the individual beliefs happen to be the same. In fact, the group should prefer the latter. The reason is that if the group has the security distribution, then there will be trading opportunities that can make everyone better off. We suspect that this preference can account for the normalization factor in the pooling procedure, a consequence of the market model presented here.

## 6  CONCLUSION

We have outlined a belief aggregation methodology based on individually rational agents (utility maximizers) trading in a competitive securities market, where the resultant equilibrium prices represent consensus beliefs. Our system is motivated from the bottom up: we begin with several common assumptions about agents (mainly that each has a subjective probability distribution, has risk averse utility for money, and behaves competitively), and seek to analyze properties of the implied price equilibrium. In this first report, we establish a few desirable properties for general markets, and provide some closed-form characterizations of demand and equilibrium for cases of constant risk aversion and limited numbers of securities. In these situations, agent demand appears intuitively rational and the price equilibrium is shown to have several appealing properties. For example, the group's aggregate demand for one security is exactly that of a single representative rational "super-agent", whose belief equals the group equilibrium price. Another significant result is the equivalence (in at least some cases) of our price equilibrium to the consensus probability generated with the normalized logarithmic opinion pool. The expert weights in the standard pool coincide with a normalized measure of risk aversion in our model, providing a decision-theoretic interpretation for an often ungrounded concept.

---

[4]One of the purposes of West's investigation was to demonstrate the pitfalls of blindly aggregating probabilities according to simple formulas (Mike West, personal communication). We share this motivation, and agree with the idea of deriving aggregate measures from behavioral postulates.

Future theoretical work will continue to generalize results to broader classes of risk-averse utility functions, and arbitrary collections of securities. We will also pursue formal characterizations concerning the existence of price equilibria, and the convergence to these equilibria via distributed bidding protocols and classical economic price adjustment. We also plan concurrent empirical investigations in more complex markets where theoretical analysis becomes intractable. Such economies may allow non-competitive agents and/or "learning" agents that update beliefs from observed prices; each extension will entail tests of existence, convergence, and properties of aggregate prices.

A natural practical application of belief aggregation is as a sub-procedure within the more general context of *group decision-making*, where agents' beliefs *and* utilities are combined to enable inference of optimal group decisions (say, choosing medical treatments). Future plans include identifying and evaluating appropriate generalizations in pursuit of a market-based approach to group decision-making in situations involving *both* asymmetric uncertainty and heterogeneous preferences.